\documentclass[letterpaper, 10 pt, conference]{ieeeconf}  

\IEEEoverridecommandlockouts                              

\usepackage{amsmath} 
\usepackage{amssymb}  

\usepackage{cite}

\usepackage{graphicx}
\usepackage{algorithm, algpseudocode}
\usepackage{amsfonts}
\usepackage{comment}
\usepackage[dvipsnames, table,xcdraw]{xcolor}
\usepackage{booktabs, tabularx}
\usepackage{tabularray}
\usepackage[normalem]{ulem}
\usepackage{hyperref}
\hypersetup{
    colorlinks=true,
    linkcolor=black,
    citecolor=black,
    urlcolor=blue,
    hidelinks 
}

\newcommand\sasa[1]{\textcolor{blue}{#1}}

\title{\Large
DPPE: Dense Pose Estimation in a Plenoxels Environment using Gradient Approximation
\vspace{-4 mm}}

\author{Christopher Kolios$^{1}$, Yeganeh Bahoo$^{1}$, and Sajad Saeedi$^{1}$
\vspace{-4 mm}
\thanks{$^{1}$
        Toronto Metropolitan University, Toronto, Canada \newline \textcolor{white}{...} emails:
        {\tt\small \{ckolios, bahoo, s.saeedi\}@torontomu.ca}}%
}

\begin{document}

\maketitle

\thispagestyle{empty}
\pagestyle{empty}

\begin{abstract}

We present DPPE, a dense pose estimation algorithm that functions over a Plenoxels environment. Recent advances in neural radiance field techniques have shown that it is a powerful tool for environment representation. More recent neural rendering algorithms have significantly improved both training duration and rendering speed. Plenoxels introduced a fully-differentiable radiance field technique that uses Plenoptic volume elements contained in voxels for rendering, offering reduced training times and better rendering accuracy, while also eliminating the neural net component. In this work, we introduce a 6-DoF monocular RGB-only pose estimation procedure for Plenoxels, which seeks to recover the ground truth camera pose after a perturbation. We employ a variation on classical template matching techniques, using stochastic gradient descent to optimize the pose by minimizing errors in re-rendering. In particular, we examine an approach that takes advantage of the rapid rendering speed of Plenoxels to numerically approximate part of the pose gradient, using a central differencing technique. We show that such methods are effective in pose estimation. Finally, we perform ablations over key components of the problem space, with a particular focus on image subsampling and Plenoxel grid resolution. \\ Project website: \href{https://sites.google.com/view/dppe}{https://sites.google.com/view/dppe}


\end{abstract}

\section{INTRODUCTION}

Camera pose estimation involves the prediction of the pose of a camera in its environment, 
with applications ranging from robotics to virtual reality headsets~\cite{Cadena2016TRO}. 
Recent developments in pose estimation involve techniques such as analysis-by-synthesis~\cite{analysisbysynthesis}, and instance-level object pose estimation~\cite{GDRNet}, which use a 3D model for training. These methods perform well, but the requirement of a detailed 3D model limits their applicability for general scenes.

In the past few years, Neural Radiance Fields (NeRFs)~\cite{NeRF} have been an active area of interest in computer vision. Given a series of images of an object from a variety of perspectives as input, once trained, a NeRF can output a rendering of the object from a previously-unseen view. 
Instead of requiring a detailed 3D model of an environment, analysis-by-synthesis-based pose estimation can use a set of images to train a NeRF, and sample from that trained NeRF. This greatly improves the generalizability of such methods. Recent works such as iNeRF~\cite{iNeRF} and BARF~\cite{BARF} have applied analysis-by-synthesis and template matching-based pose estimation to NeRF scenes, with strong performance. However, the long training time and slow image rendering speed of classical NeRF hampers the performance of these methods.


Several NeRF-based works have improved upon these limitations. Some of the most influential among these works include Plenoxels~\cite{Plenoxels}, Instant Neural Graphics Primitives (Instant NGP)~\cite{instantNGP}, and 3D Gaussian Splatting~\cite{kerbl3Dgaussians}. While each method uses a different mechanism, they all significantly reduce training time, as well as the time it takes to render novel views. In particular, Plenoxels discretizes NeRF, storing a voxel grid of spherical harmonic coefficients and densities, which can be trilinearly interpolated to represent a continuous scene. Plenoxels grid ``training" is over 100$\times$ faster than that of NeRF, providing more photorealistic scene representations, and quicker rendering as well.
\begin{figure}[t!]
    \centering
    {\includegraphics[width=0.9\columnwidth]{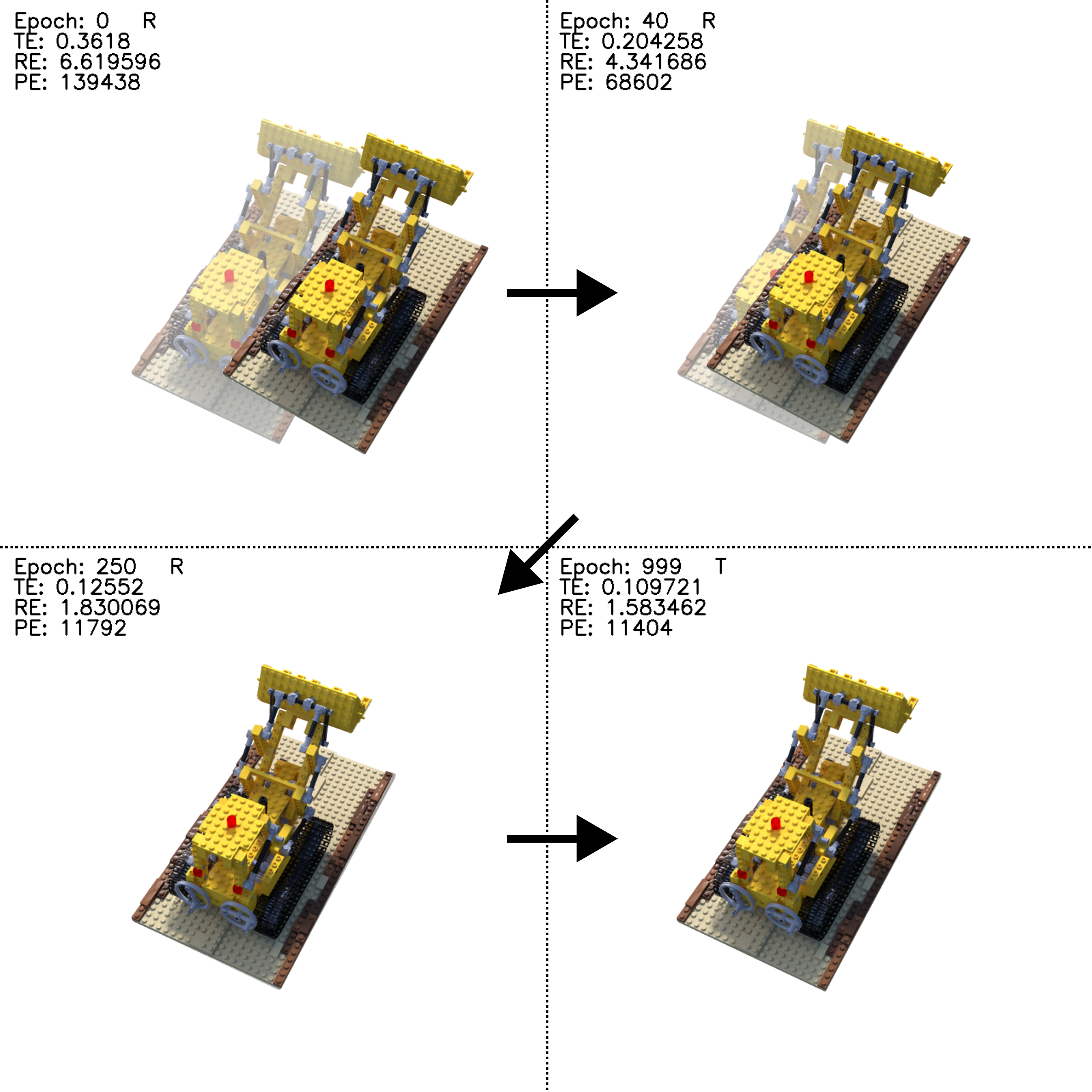}} 
    \vspace{-4 mm}
    \caption[DPPE Qualitative Performance]{Qualitative performance of DPPE on a pose from the Lego scene, rendered from the current pose estimate (opaque) with the ground-truth image (transparent). DPPE starts with the initial perturbed pose (top left), then as it runs the pose estimate gets closer to the ground truth (left to right, top to bottom).}
    \label{fig:dppe_qualitative}
\vspace{-4 mm}
\end{figure}
Inspired by iNeRF~\cite{iNeRF}, and BARF~\cite{BARF}, in this paper, we explore the extension of classical pose estimation techniques to a Plenoxels environment, which we present as Dense Pose estimation in a Plenoxels Environment (DPPE) (See Fig.~\ref{fig:dppe_qualitative}). 
While iNeRF is successful in its pose estimation, the slow rendering time of NeRF means that only a subset of rays from each image can be rendered at each pose estimation step, for termination in a reasonable time frame. In contrast, pose estimation algorithms operating over a Plenoxels grid can re-render the full image at each step, while still maintaining reasonable runtimes. Plenoxels scene rendering is fast enough that more than 6 image renders can be performed in the time that it would take a single NeRF~\cite{NeRF} render. With this in mind, traditional approaches from computer graphics that use central differencing to estimate image gradients~\cite{BeyondTrilerp, prattimageprocessing} can be run over all 6 camera DoF. This image gradient can be used in combination with template matching~\cite{lucaskanade, BARF} to perform pose estimation.

In our approach, the photometric error between the rendered image and the ground-truth image is computed for each epoch and used with the image gradient approximation to take the camera pose ``step" that minimizes that error. This iterative process is repeated until the error is sufficiently small or the epoch max is reached. Similar to NeRF, Plenoxels allows for a subset of image rays to be rendered. Following iNeRF, we perform ablations on the effect of random ray subsampling on pose estimation performance and algorithm runtime, and demonstrate the tradeoffs in processing the entire image versus a subset of the image. We also examine the effect of the Plenoxels grid resolution in this context.

To summarize, this work makes the following contributions:
    (1) We show that DPPE can perform 6 DoF camera pose estimation, using only a monocular RGB image, a pre-trained Plenoxels grid, and a perturbed pose as input.
    (2) We provide comparisons to existing work, and quantitatively and qualitatively analyze both the advantages and limitations of the proposed method, and
    (3) We evaluate the effectiveness of DPPE on standard test scenes, performing ablations on key algorithm characteristics such as grid resolution, and percent of image sampled. 

This paper is organized as follows: Sec.~\ref{sec:related_wrk} reviews relevant literature. Sec.~\ref{sec:background} presents the background. The proposed method is detailed in Sec.~\ref{sec:method}. Experimental results are presented in Sec.~\ref{sec:results}. Finally, Sec.~\ref{sec:conclusion} concludes the paper.

\section{Literature Review}\label{sec:related_wrk}

\textbf{Neural Rendering.} NeRF~\cite{NeRF} uses a neural network to represent a scene in its entirety. To accomplish this, NeRF uses a multilayer perceptron (MLP) network, where the function to be optimized is one that takes in a single continuous 5D coordinate (consisting of spatial location ($x, y, z$) and viewing direction ($\theta$, $\phi$)), and the output is volume density and view-dependent emitted radiance (colour) at that location. The weights and biases of the NeRF MLP are able to represent the entire scene, after training on images from that scene. As input, NeRF uses images of an object or scene from a variety of known perspectives. Techniques such as positional encoding~\cite{instantNGP, nerffourier} and two-pass rendering have been used to improve results. Many works have focused on improving upon aspects of NeRF, from render time~\cite{Plenoxels, instantNGP, kerbl3Dgaussians}, to scale~\cite{mipnerf, blockNeRF} , to extension to video~\cite{neuralflow}.

One of the more unique algorithms that improved upon NeRF was Plenoxels~\cite{Plenoxels}, which uses a 3D voxel grid of Plenoptic volume elements to represent an environment. The corner of each grid voxel contains a set of spherical harmonic coefficients and densities. Using a slightly-modified rendering algorithm, a ray is cast into the scene, with the resulting colour of the ray being dependent on the accumulation of colours from points sampled through the voxel grid. The spherical harmonics are trilinearly interpolated for raycasts. This process is fully-differentiable end-to-end, allowing for efficient optimization. Given input images as well as their respective camera poses, photometric loss is computed along each ray and used to optimize the spherical harmonic coefficients and densities. Very recent works such as 3D Gaussian Splatting~\cite{kerbl3Dgaussians} use spherical harmonics in their scene representation as well, with the addition of Gaussians.

\begin{figure*}[t!]
    \centering
    {\includegraphics[width=.9\textwidth]{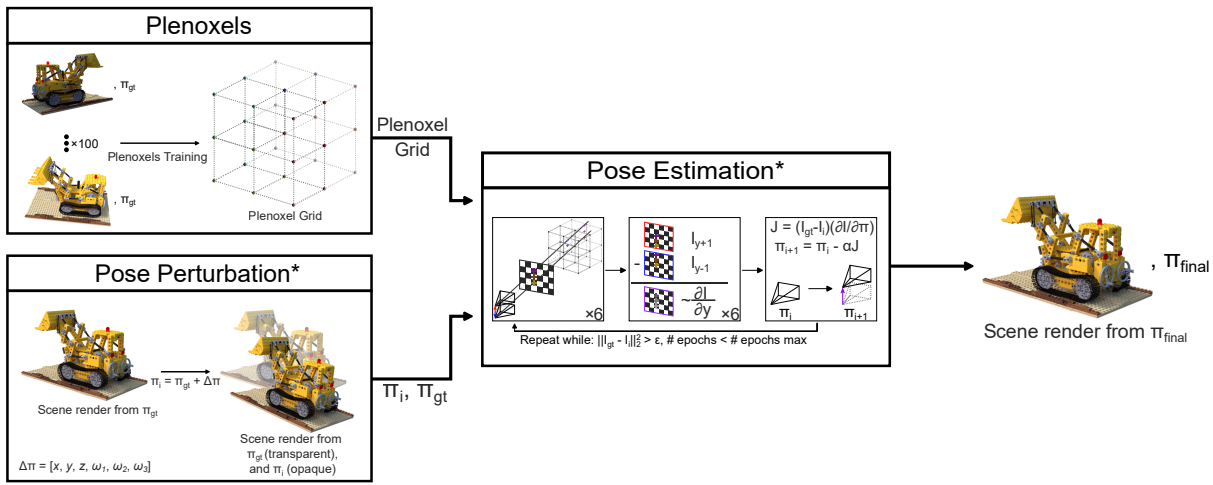}} 
    \caption[Analysis Pipeline]{An illustration of the analysis pipeline. A star indicates a significant contribution of this work towards the module. First, a Plenoxels grid is trained for a scene. Then, a ground-truth pose is perturbed, with the extent of perturbation depending on the test being run. The trained Plenoxels grid, ground-truth pose, and perturbed pose are passed into the pose estimation process, which outputs the final pose after optimization. The scene can be rendered from any pose, to visualize camera position and orientation.}
    \label{fig:pipeline}
\end{figure*}

\textbf{Classical Pose Estimation.} Pose estimation 
is a classic and long-studied problem in computer vision. Traditional pose estimation techniques typically fall into one of two categories: sparse, or dense. Sparse methods look to form correspondences between sparse sets of landmarks representing key features within an environment. These key features are often image textural features, such as SIFT or ORB, which represent the variations of contrast within image patches and are typically stored in sparse point clouds. The formulation of these correspondences to recover camera pose is called the Perspective-n-Point problem~\cite{pnp}, for which a variety of common solutions exist, such as RANSAC PnP. One of the most widely-used Structure from Motion (SfM) algorithms, COLMAP~\cite{colmap}, uses PnP as part of its pipeline. Mono-SLAM~\cite{monoslam}, ORB-SLAM3~\cite{orbslam3}, and other SLAM algorithms~\cite{siftslam} also use sparse features as part of their localization procedure. Dense pose estimation approaches, on the other hand, look to optimize pose by matching within complete scene representations. Classical approaches to this problem include template matching, and the Lucas-Kanade Method~\cite{lucaskanade}. Some SLAM algorithms, such as KinectFusion~\cite{kinectfusion}, or $\nabla$-SLAM~\cite{gradSLAM} use such representations. Finally, researchers have recently been investigating the use of deep learning for pose estimation. For example, convolutional neural networks have been used to estimate optical flow for the purpose of dense pose estimation, with good results~\cite{cNNopticalflow}.

\textbf{Neural Pose Estimation.} Very recently, NeRF-based scenes have been used as the environment on which pose estimation is performed. Recent radiance field-based methods for pose estimation can involve using monocular RGB camera data\cite{iNeRF, NeRF--, BARF, parallel_inv, locnerf, monoslam}, RGB + Depth (RGB-D) data ~\cite{iMAP, NICE-SLAM, plenoxelslam, kinectfusion, orbslam3}, or even LiDAR~\cite{LiDARSLAM}. The most widely applicable among these methods are those that use a singular RGB camera, due to the low cost and ubiquity of regular cameras. There are sparse feature-matching approaches that have been used with neural environments, such as Orbeez-SLAM~\cite{orbeezslam}, as well as dense template matching approaches, such as BARF~\cite{BARF}. Notable works in the context of SLAM + NeRF include iMAP~\cite{iMAP}, NICE-SLAM~\cite{NICE-SLAM}, and NeRF-SLAM~\cite{NeRF-SLAM}. While these SLAM works are effective at localization, with the exception of NeRF-SLAM, they rely on RGB-D data, which is not always available. Many recent works that focus particularly on pose estimation use an analysis-by-synthesis approach to take advantage of the useful re-rendering capability of NeRFs. These works include iNeRF~\cite{iNeRF}, NeRF$- -$~\cite{NeRF--}, BARF~\cite{BARF}, and Parallel Inversion of NeRFs~\cite{parallel_inv} (the latter using Instant NGP for pose estimation).

The general framework of such algorithms involves using a pre-trained neural scene~\cite{iNeRF, parallel_inv}, or doing on-the-fly training~\cite{NeRF--, BARF} of a NeRF. Given a pose estimate, a subset of rays from the image plane is cast into the scene and the network is evaluated, resulting in a set of rendered pixels. These rendered pixels on the image plane are then compared with the observed pixels from the ground-truth image, and a loss term is calculated. This loss is then backpropagated, and used to update the current pose estimate. This process is repeated until the image from the estimate matches with the ground truth image, or the algorithm times out. Each work contributed something novel to the process - iNeRF~\cite{iNeRF} investigated the influence of different ray sampling techniques, such as random sampling and interest-region sampling, and Parallel Inversion of NeRFs~\cite{parallel_inv} used a Monte-Carlo-based sampling to help overcome the problem of local minima. While there has been plenty of work on NeRF, investigation on Plenoxels is very limited - only a single work has investigated this~\cite{plenoxelslam}. 

Teigen et al. introduced 
a method that performs RGB-D SLAM on an environment, using Plenoxels for both mapping and tracking~\cite{plenoxelslam}. 
Their tracking formulation is similar to that of BARF~\cite{BARF}, using an analytic derivative from the trilinear interpolation to minimize photometric loss as a function of pose. Our work approaches pose estimation differently in two key areas. Firstly, we solely use monocular RGB data, whereas \cite{plenoxelslam} requires depth data for their algorithm to function. While using depth data for mapping clearly can improve performance and assist in regions where there is little colour gradient information, it reduces the applicability and accessibility of the approach - depth cameras are not always available. Secondly, our work uses a numerical approximation of the derivative for the image gradient, as opposed to the analytically-derived gradient in \cite{plenoxelslam}. While the analytical derivative can be intimately integrated with the Plenoxels CUDA code for performance speedup, the result is that the method is entangled with the specific environment representation (Plenoxels). In contrast, DPPE, while here implemented over a Plenoxels grid, could in theory be used in any kind of setting in which re-renderings are not prohibitively expensive, as the numerical gradient approximation technique makes no assumptions regarding the underlying scene representation.

\section{Background}\label{sec:background}

\textbf{Plenoxels.} From Plenoxels~\cite{Plenoxels}, the colour $\hat{C}$ of a rendered ray $\mathbf{r}$ is given by:
\begin{equation} \label{eq:plenoxelrender}
    \hat{C}(\mathbf{r}) = \sum_{i=1}^{N} T_i(1-\exp(-\sigma_i\delta_i))\mathbf{c_i} \ ,
\end{equation}

such that $\sigma_i$ indicates the opacity of sample $i$ out of $N$ total samples, $\delta_i$ indicates the distance between sample $i$ and sample $i+1$, $\mathbf{c_i}$ indicates the colour of sample $i$ as determined by trilinear interpolation of the 8 nearest voxels, and:
\begin{equation} \label{eq:plenoxelt}
    T_i = \exp(-\sum_{j=1}^{i-1}\sigma_j\delta_j) \ ,
\end{equation}

denotes how much light is transmitted through ray $\mathbf{r}$ to sample $i$ (as opposed to being contributed by preceding samples). See:~\cite{Plenoxels, Plenoctrees} for more details.




\textbf{BARF's Formulation.} BARF~\cite{BARF} seeks to optimize both camera pose and NeRF weights via a bundle-adjustment process. They set up their minimization problem as follows:
\begin{equation} \label{eq:barf}
    \min_{\pi_1, ... \pi_m, \theta}\sum_{i=1}^{M}\sum_{\mathbf{p}}||\hat{I}(\mathbf{p} = (u, v), \pi_i, \theta) - I(\mathbf{p})||_2^2 \ ,
\end{equation}
where $\pi_1, ... \pi_m$ are the series of camera poses, $\theta$ is the NeRF network weights, $M$ is the number of images, $\textbf{p} = (u, v)$ is a vector of pixels in the image, and $I$ and $\hat{I}$ are the ground truth and scene-rendered images, respectively.

\section{METHODOLOGY} \label{sec:method}
In this section, pose representation, the problem setup, and rendering are explained. Following this, the proposed pose optimization technique is presented. Fig.~\ref{fig:pipeline} shows the outline of the proposed method. A Plenoxels grid is trained in advance, from which a pose is selected. This pose is then passed in to DPPE after being perturbed, with the goal of recovering that original pose.

\subsection{Camera Pose Representation}

Camera pose can be represented as some $\pi \in SE(3)$ such that it can be decomposed into: $\pi = [\pi_R|\pi_\mathbf{t}]$ where $\pi_R \in SO(3)$ is the camera rotation, and $\pi_\mathbf{t} \in \mathbb{R}^3$ is the camera translation. $SE(3)$ is the Special Euclidean Lie Group, and $SO(3)$ is the Special Orthogonal Lie Group. They are ultimately $4\times4$ and $3\times3$ matrices which represent the differentiable manifolds of 6 degrees of freedom (rotation + translation) or 3 degrees of freedom (rotation only) for cameras, respectively. The tangent plane at the identity element for these manifolds is called the Lie Algebra, which has an associated Cartesian representation. 
For $SO(3)$, the Cartesian representation can be parameterized as:
\begin{equation} \label{eq:cartesian}
    \boldsymbol{\omega} = (\omega_1, \omega_2, \omega_3) \in \mathbb{R}^3 \ ,
\end{equation}
which represents the 3 degrees of freedom of rotation. 
The Lie Algebra will be:
\begin{equation} \label{eq:lie}
    \boldsymbol{\omega} \hat{} = [\boldsymbol{\omega}]_\times = \begin{bmatrix}
0 & -\omega_3 & \omega_2\\
\omega_3 & 0 & -\omega_1\\
-\omega_2 & \omega_1 & 0
\end{bmatrix} = \theta \mathbf{u} \ ,
\end{equation}
such that $\mathbf{u} = (x, y, z)$ is a unit vector representing an axis of rotation, and $\theta$ is an angle with which to rotate about that axis. Hat ($\ \hat{} \ $) and vee (${}^\vee$) will move a rotation from the Cartesian to Lie Algebra representation, and the Lie Algebra to Cartesian representation, respectively~\cite{microLie}. For convenience, we represent our camera poses using a full $SE(3)$ camera-to-world matrix, and convert the rotation matrix component to axis-angle or Cartesian representation via hat and vee as needed for pose optimization steps.

\subsection{Problem Setup}

Similar to iNeRF~\cite{iNeRF} and part of BARF~\cite{BARF}, our goal is to invert a scene representation to recover the ground-truth camera pose. In the case of DPPE, the scene is represented by a pre-trained Plenoxels grid, which can be rendered from via Eq.~\eqref{eq:plenoxelrender}. Therefore, the available inputs are the pre-trained Plenoxels grid, the image of a scene $I_{gt}$ from a ground-truth pose $\pi_{gt}$, and an initial pose estimate $\pi_{i}$, which is a camera pose obtained by perturbing the ground truth pose. The goal is to output a camera pose $\pi_{final}$ that is as close to $\pi_{gt}$ as possible, via the minimization of photometric error between the rendered image $\hat{I}$ from the camera pose estimate $\pi_i$, and the ground truth image $I_{gt}$, for a set of pixels $\textbf{p}$. The optimization problem we are trying to solve is therefore:
\begin{equation} \label{eq:minimization}
    \pi_{final} = \min_{\pi_i} \sum_{\textbf{p}} ||\hat{I}(\mathbf{p}, \pi_i) - I_{gt}(\mathbf{p})||_2^2 \ .
\end{equation}

\subsection{Rendering}

The full image can be used, in which case $|\textbf{p}| = W \times H$, where $W$ and $H$ are the width and the height of the image, respectively, or a subset of pixels from the image can be used, such that $\textbf{p} \subset I$. As image pixels are formed by projecting rays from the camera onto an image plane and rendering from a scene, the ray associated with a pixel $\textbf{p}$ with image coordinates $(u, v)$, assuming a camera pose $\pi = [\pi_R |\pi_t]$, is given by: $\textbf{r} = \textbf{o}_\textbf{r} + h \textbf{d}_\textbf{r}$. Here, $h \in \mathbb{R}$ is the length of the ray, $\textbf{o}_\textbf{r} = \textbf{o}_{\pi} = \pi_t$ is the origin of the ray, which starts at the camera's origin, and,
\begin{equation}
    \textbf{d}_r = \pi_R
    \begin{bmatrix}
        \frac{u-\frac{W}{2}}{f} \\
        -\frac{v-\frac{H}{2}}{f} \\
        -1
    \end{bmatrix} \ ,
\end{equation}
gives the direction of the ray, assuming a camera focal length $f$. Thus, the rendering can be obtained by casting the rays $\textbf{r} \ \forall$ pixels $\textbf{p}$ into the Plenoxels grid, such that the rendered pixels $\hat{I}(\textbf{p}, \pi_i) = \hat{C}(\textbf{r})$, by Eq.~\eqref{eq:plenoxelrender}.

\subsection{Optimization}

To find the $\pi$ that minimizes the photometric error between the rendered and ground-truth images (Eq.~(\ref{eq:minimization})), optimization must be done. Here, we adopt a similar optimization technique to that of BARF~\cite{BARF} and classical template matching~\cite{lucaskanade}, using the ``steepest descent image" defined by:
\begin{equation} \label{eq:ppe_sdi}
    \mathbf{J}(\mathbf{p}, \pi) = \sum_{i=1}^{N} \frac{\partial\hat{C}(V_1, ... V_N)}{\partial V_i} \frac{\partial V_i (\pi)}{\partial x_i (\pi)} \frac{\partial W(x_i; \pi)}{\partial \pi} \ ,
\end{equation}
such that $V_i = [\sigma_i; \mathbf{c}_i]$ is a voxel containing a density $\sigma$ and colour $\mathbf{c}$, $x_i$ is a spatial location, and $W$ is a transformation from camera view space to 3D world coordinates, for $N$ pixels. Ultimately, this $\frac{\partial C}{\partial \pi}$ represents the change in a rendered pixel colour given a change in camera pose. While this could be analytically derived (see~\cite{plenoxelslam}), this would tie the pose estimation technique to Plenoxels specifically.

\begin{algorithm}[b!] 
\caption{Render-Based Plenoxels Pose Estimation}\label{alg:render_ppe}
\begin{algorithmic}[1]
    \While{$||\hat{I}_{\pi_c}-I ||_{2}^{2}> \epsilon$} \label{al:conv}
    \State $\mathbf{r} \gets$ set of rays to be cast from current pose $\pi_c$ \label{al:r}
    \State $\hat{I}_{\pi_c}(\mathbf{r}) \gets$ \Call{VolumeRender}{$\mathbf{r}$} \label{al:rr} 
    \ForAll{$i \in [x,y,z,\omega_1,\omega_2,\omega_3]$} 
        \State $\hat{I}_{\pi_c}(R_{(i+1)}) \gets$ \Call{VolumeRender}{$R_{(i+1)}$}
        \State $\hat{I}_{\pi_c}(R_{(i-1)}) \gets$ \Call{VolumeRender}{$R_{(i-1)}$}
        \State $J_i(\mathbf{r}) \gets [\hat{I}_{\pi_c}(R_{(i+1)}) - \hat{I}_{\pi_c}(R_{(i-1)})]/2$ \label{al:j} 
    \EndFor
    \State $\Delta I(\mathbf{r}) \gets \hat{I}_{\pi_c}(\mathbf{r})-I(\mathbf{r})$ \label{al:di}
    \State $SGD: \pi_c \gets \pi_c + \alpha (J(\mathbf{r}) \cdot \Delta I(\mathbf{r})) $ \label{al:sgd} 
    \EndWhile
\end{algorithmic}
\end{algorithm}

For a more generalizable pose optimization, a numerical approximation of Eq.~\eqref{eq:ppe_sdi} can be determined by comparing the change in image from $\pi_c \pm 1$ for all 6 camera DoF $[x,y,z,\omega_1,\omega_2,\omega_3]$, and estimating the gradient via central differencing. For example, $\frac{\partial C}{\partial \pi_x} \approx [C(\pi_{x + 1}) + C(\pi_{x - 1})]/2$. This method of image gradient estimation is commonly used in computer graphics applications~\cite{BeyondTrilerp}. As Plenoxels rendering uses CUDA, this re-rendering is faster than other NeRF-based approaches. Independent of how Eq.~\eqref{eq:ppe_sdi} is computed, it can be used with stochastic gradient descent to compute a pose update step via:
\begin{equation} \label{eq:SGD}
    \pi_c \gets \pi_c + \alpha (J(\mathbf{r}) \cdot \Delta I(\mathbf{r})) \ ,
\end{equation}
where $\pi_c$ is the current pose estimate, $\mathbf{r}$ is the set of rays to render, $J(\mathbf{r})$ is the steepest descent image, $\Delta I(\mathbf{r})$ is the photometric error between the rendered image and the ground truth image for the set of rays, and $\alpha$ is a learning rate parameter. 
Alg.~\ref{alg:render_ppe} summarizes the proposed method. It runs until convergence (line~\ref{al:conv}).
Variable $\mathbf{r}$ can either be the rays forming a full image, or a subset of rays (line~\ref{al:r}). For all 6 camera DoF, the scene is re-rendered from $\pm \ 1$ ``step'' (loop on line~\ref{al:rr}), and used to form a numerical approximation of the image gradient via central differencing (line~\ref{al:j}). Finally, this is used alongside the photometric error of the rays (line~\ref{al:di}) to perform an SGD step and update the pose estimate (line~\ref{al:sgd}). It should be noted that the $\pm 1$ for re-rendering in the translational case describes a shift in voxel, which has a variable size depending on the Plenoxel grid resolution. The grid resolution (which is typically a power of $2$) is a customizable parameter that can be dynamically upsized or downsized when training the Plenoxel grid, such that higher resolution results in better reconstruction quality at the cost of compute time and memory, and vice-versa~\cite{Plenoxels}. For rotation, each re-rendering is conducted with a step size of $0.001$ radians. For each epoch, either a translation step or a rotation step is performed, updating the pose estimate's corresponding 3 DoF. The optimization strategy consists of alternation between a single rotation step and a single translation step until the maximum epoch limit is reached or photometric error drops below an error threshold $\epsilon$.

\begin{table}[t]
\caption{Pose estimation error for iNeRF~\cite{iNeRF} and DPPE on the NeRF Synthetic Dataset. Per-dataset results are presented as the average over 20 perturbations. -Full indicates that the full image was used for pose estimation (640,000 pixels), and -1\% indicates that 1\% of the image was used (6,400 pixels).}
\label{tab:my-table}
\setlength{\tabcolsep}{1.2pt}
\begin{tabularx}{\columnwidth}{@{}llcccccccccc@{}}
\toprule
 &  & Chair & Drums & Ficus & Hotdog & Lego & Materials & Mic & Ship &  & Average \\ \cmidrule(lr){3-10} \cmidrule(l){12-12} 
Method &  & \multicolumn{10}{c}{\% of Rotation Errors \textless  5° (↑)} \\ \cmidrule(r){1-1}
iNeRF &  & 0.8 & 0.75 & 0.6 & \textbf{1} & 0.95 & 0.7 & 0.55 & \textbf{0.9} &  & 0.78 \\
DPPE-full &  & \textbf{1} & \textbf{0.95} & 0.7 & 0.75 & \textbf{1} & \textbf{0.8} & \textbf{0.85} & 0.85 &  & \textbf{0.86} \\
DPPE-1\% &  & \textbf{1} & 0.85 & \textbf{0.95} & 0.9 & 0.95 & 0.65 & 0.8 & 0.8 &  & \textbf{0.86} \\ \cmidrule(l){3-12} 
 &  & \multicolumn{10}{c}{\% of Translation Errors \textless  0.2 units (10\% of width) (↑)} \\
iNeRF &  & \textbf{0.7} & 0.65 & 0.55 & \textbf{1} & \textbf{0.95} & \textbf{0.65} & 0.55 & \textbf{0.9} &  & \textbf{0.74} \\
DPPE-full &  & \textbf{0.7} & 0.65 & 0.55 & 0.35 & 0.8 & 0.5 & \textbf{0.6} & 0.35 &  & 0.56 \\
DPPE-1\% &  & 0.55 & 0.65 & \textbf{0.65} & 0.45 & 0.8 & 0.55 & 0.55 & 0.4 &  & 0.58 \\ \cmidrule(l){3-12} 
 &  & \multicolumn{10}{c}{Average Rotation Error (°) (↓)} \\
iNeRF &  & 4.95 & 3.71 & 6.30 & \textbf{0.30} & \textbf{0.59} & 4.85 & 6.59 & \textbf{1.28} &  & 3.57 \\
DPPE-full &  & \textbf{2.43} & \textbf{2.36} & 3.55 & 3.98 & 1.74 & \textbf{3.19} & \textbf{2.44} & 3.92 &  & 2.95 \\
DPPE-1\% &  & 2.48 & 2.61 & \textbf{2.36} & 3.46 & 1.89 & 3.83 & 2.98 & 3.67 &  & \textbf{2.91} \\ \cmidrule(l){3-12} 
 &  & \multicolumn{10}{c}{Average Translation Error (units)   (↓)} \\
iNeRF &  & 0.31 & 0.31 & 0.45 & \textbf{0.02} & \textbf{0.03} & 0.37 & 0.32 & \textbf{0.08} &  & 0.24 \\
DPPE-full &  & \textbf{0.17} & \textbf{0.17} & 0.21 & 0.27 & 0.12 & 0.23 & 0.22 & 0.27 &  & 0.21 \\
DPPE-1\% &  & 0.18 & 0.18 & \textbf{0.17} & 0.24 & 0.13 & \textbf{0.21} & \textbf{0.21} & 0.26 &  & \textbf{0.20} \\ \cmidrule(l){3-12} 
 &  & \multicolumn{10}{c}{Average Time Elapsed (seconds) (↓)} \\
iNeRF &  & 13.3 & 13.4 & 12.8 & 13.3 & 13.2 & 13.2 & 11.8 & 13.3 &  & 13.0 \\
DPPE-full &  & 63.8 & 62.5 & 61.6 & 77.4 & 70.2 & 66.8 & 62.4 & 91.9 &  & 69.6 \\
DPPE-1\% &  & \textbf{10.7} & \textbf{10.6} & \textbf{10.6} & \textbf{11.0} & \textbf{10.9} & \textbf{10.7} & \textbf{10.5} & \textbf{11.5} &  & \textbf{10.8} \\ \bottomrule
\end{tabularx}
\end{table}
\section{RESULTS AND DISCUSSION}\label{sec:results}

In this section, we begin with an evaluation of the performance of DPPE over the NeRF synthetic dataset~\cite{NeRF}. Additionally, we perform comparisons to iNeRF. This is followed by an ablation over key algorithm parameters, including pixel sub-sampling and Plenoxel grid resolution. Finally, the limitations of the work are considered.

\subsection{Implementation Details}

All experiments were run on a nVIDIA RTX-4090 GPU (24 GB VRAM), with a maximum cutoff of 1000 epochs. Pytorch was used for handling most of the data, with a CUDA backend for the SGD step and the Plenoxels rendering. The full results generation pipeline is presented in Fig.~\ref{fig:pipeline}. The photometric error cutoff $\epsilon$ was set to 2000 (Alg.~\ref{alg:render_ppe}-line~\ref{al:conv}), except for when using 1\% of the image, where it was set to 0. The learning rate $\alpha$ for the translation is initialized as $4$ and decayed every 100 epochs by $25\%$, and the rotation learning rate is initialized at $5$, with the same relative rate of decay. These values were manually tuned, but are consistent across all datasets and tests.


\subsection{Synthetic Dataset} \label{sec:synth}

\textbf{Setting:} The NeRF synthetic dataset~\cite{NeRF} contains 8 synthetic scenes, with $800 \times 800$ images captured from a full $360$\textdegree \ around an object, simulated via Blender~\cite{Blender}. Some of the objects are realistic simulations of non-Lambertian materials, including reflections and transparency. The object scenes are known as: Chair, Drums, Ficus, Hotdog, Lego, Materials, Mic, and Ship. They offer varying levels of complexity, and each presents a unique challenge in the context of pose estimation. Exact camera poses, including camera intrinsics, are stored. There are 100 training images, 100 validation images, and 200 test images, with known camera poses. NeRF-synthetic was used in the first NeRF paper~\cite{NeRF}, and is commonly used in neural pose estimation works, including iNeRF~\cite{iNeRF}, BARF~\cite{BARF}, and more~\cite{parallel_inv}.

Joint translation and rotation pose estimation was performed with an initial perturbation of up to $80$ pixels (0.2 units) in each spatial axis, and $5$\textdegree \ in each rotational ``axis''. $80$ pixels is $10\%$ of the width and height of the image. Perturbation was done by starting with the ground truth pose, then randomly generating perturbations up to the aforementioned magnitudes, and randomly adding or subtracting those perturbations to each of the 6 camera DoF.

Rotational error (RE) and translation error (TE) are used as the evaluation criteria. These metrics are computed for a sample of 20 different poses (randomly-selected from the validation set) for all scenes for more robust results. A rotational success cutoff of RE $= 5$\textdegree \ (the same as in iNeRF~\cite{iNeRF}, Parallel Inversions~\cite{parallel_inv}) is used. The translational success cutoff is set to $80$ pixels (vs. 5 cm for iNeRF). This cutoff corresponds to $10$\% of the image side length, and is selected as it represents a minimal translational difference expressed relative to a known image quantity. Real-world distances were not available for the selected dataset.

The Plenoxels grid was trained using the default configurations from Plenoxels~\cite{Plenoxels}, such that the grid resolution was $[256]^3$. For comparisons to iNeRF, a NeRF was trained from scratch for the synthetic scenes. These were trained using the default configuration files from NeRF~\cite{NeRF}, up to 200K iterations. It should be emphasized that a completely one-to-one comparison between iNeRF~\cite{iNeRF} and our method (DPPE) cannot be made, as they function over different scene representation techniques (NeRF~\cite{NeRF} and Plenoxels~\cite{Plenoxels}, respectively). However, as they use the same error metrics and have the same goal, it is useful to consider their relative performance. To form a more fair comparison, iNeRF's random pixel selection strategy was used. The batch size for iNeRF was set to 512 pixels, with full image resolution. Given the similar runtimes, iNeRF is most comparable to DPPE using 1\% of the image (6,400 pixels).

\begin{figure*}[ht]
    \centering
    {\includegraphics[width=0.85\textwidth]{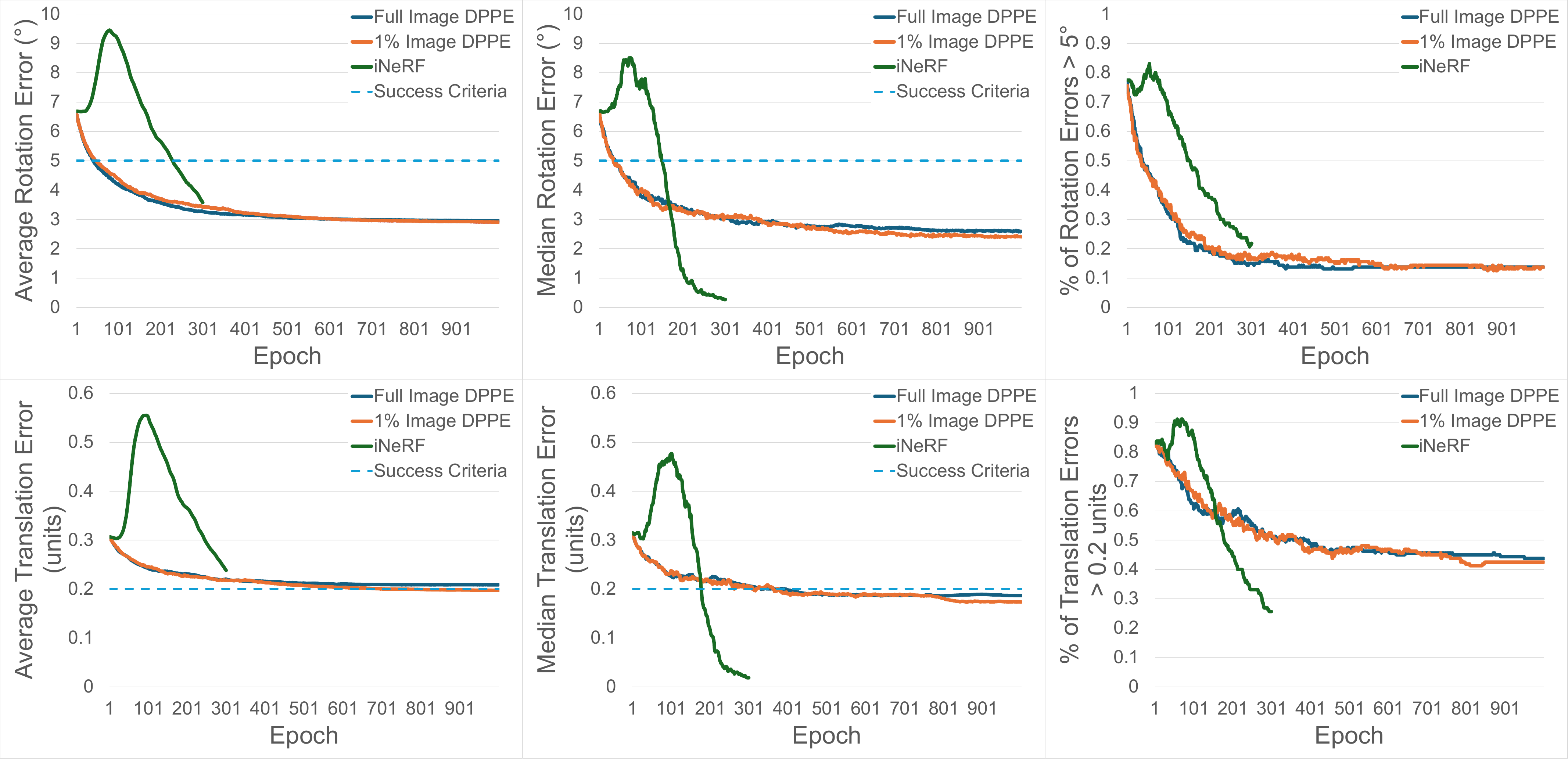}} 
    \caption[DPPE Performance Over Time]{Rotation (top) and translation (bottom) pose error for DPPE. From left to right, pose error is presented as: average, median, and \% failures, as a function of epoch. DPPE quickly converges towards the success criteria before tapering off, whereas iNeRF starts by getting worse, before rapidly converging to a very low error.}
    \label{fig:dppe_perf_time}
\end{figure*}

\textbf{Results:} A summary of the results is presented in Table~\ref{tab:my-table}. While the comparison is not one-to-one due to the different scene representation techniques, as the evaluation criteria, initial perturbations, start goal, and end goal are the same, it is still worthwhile to compare performance. For all scenes except Hotdog, Lego, and Ship, DPPE had a lower average rotation and translation error. On average, for rotations, DPPE had better performance than iNeRF. Translationally, while iNeRF had a greater average percent of pose estimations that met the error cutoff, iNeRF's average translation error was higher, due to worse performance translationally in all scenes except Hotdog, Lego, and Ship. DPPE using the full image for pose estimation took approximately 70 seconds for full execution, whereas when using only 1\% of the image the average runtime decreased to 11 seconds. DPPE-1\% was faster than iNeRF, even though its batch size was larger, mostly due to Plenoxels rendering being quicker. For DPPE, Hotdog, Materials, and Ship were some of the most challenging datasets. As DPPE is reliant on the image gradient, for scenes with large regions of uniform texture (e.g. the bun of Hotdog, the water of Ship), the algorithm struggles. Due to the use of SGD, DPPE can sometimes get stuck for scenes with local minima (e.g. the grid of Materials), depending on pose initialization. Pose error as a function of epoch is shown in Fig.~\ref{fig:dppe_perf_time}. DPPE tends to immediately begin with convergence towards the goal, whereas iNeRF gets worse before it improves. DPPE tends to fail more gracefully than iNeRF - getting stuck in a poor local minimum instead of continually getting worse. This can be observed from the average vs. median relative performances. Additionally, while the rapid initial convergence of DPPE is a desirable property (e.g. for integration with SLAM), it is unable to reach the small errors that iNeRF sometimes finds. Part of the reason for why this may be is due to the $\pm \ 1$ sampling step used for image gradient estimation. Since the size of each step is kept constant throughout runtime, once the algorithm reaches a state in which the error is smaller than the resampling step, the gradient approximation becomes too coarse to capture the information. Future work could investigate the impact of decreasing the step size as a function of epoch. Interestingly, the performance of DPPE-1\% is on-par with DPPE-full. Motivated by this, we investigate the effect of ray subsampling in the following section.

\begin{table}[]
\caption{DPPE results with variable ray subsampling. \%Im indicates the percent of pixels from the image randomly selected for pose estimation. \%RE $< 5$\textdegree and \% TE $< 0.2$ units indicate the percent of rotation and translation errors that meet the success criteria.}
\label{tab:table2}
\setlength{\tabcolsep}{4pt}
\begin{tabular}{@{}llccccc@{}}
\toprule
 &  & \%RE \textless 5° & \%TE \textless 0.2 units& Avg. RE & Avg. TE& Runtime \\
Im\% &  &  (↑)             & (↑)                       &   (°) (↓)      &  (units) (↓)           &  (sec) (↓) \\\cmidrule(r){1-1} \cmidrule(l){3-7} 
0.01 &  & 0.85 & 0.575 & 2.97 & 0.20 & \textbf{11.0} \\
0.25 &  & 0.875 & 0.55 & 3.15 & 0.22 & 23.0 \\
0.50 &  & 0.85 & \textbf{0.65} & 2.94 & \textbf{0.19} & 45.4 \\
0.75 &  & 0.86 & 0.56 & 2.96 & 0.21 & 66.8 \\
1.00 &  & \textbf{0.91} & 0.59 & \textbf{2.82} & 0.20 & 73.1 \\ \bottomrule
\end{tabular}
\end{table}

\begin{table}[]
\caption{DPPE results with variable Plenoxel grid resolution. \%RE $< 5$\textdegree and \%TE $< 0.2$ units indicate the percent of rotation and translation errors that meet the success criteria.}
\label{tab:table3}
\setlength{\tabcolsep}{3pt}
\begin{tabularx}{\columnwidth}{@{}ccccccc@{}}
\toprule
 &  & \%RE \textless 5°   & \%TE \textless 0.2 units & Avg. RE  & Avg. TE     & Runtime  \\ 
Resolution &  & (↑)                 &              (↑) &  (°) (↓) & (units) (↓) & (sec) (↓)\\\cmidrule(r){1-1} \cmidrule(l){3-7} 
64 &  & 0.31 & 0.18 & 7.98 & 0.56 & \textbf{33.1} \\
128 &  & 0.56 & 0.31 & 5.06 & 0.36 & 47.9 \\
256 &  & \textbf{0.91} & 0.59 & \textbf{2.82} & 0.20 & 73.1 \\
512 &  & 0.81 & \textbf{0.69} & 2.85 & \textbf{0.18} & 77.8 \\ \bottomrule
\end{tabularx}
\end{table}
\subsection{Pixel Sub-Sampling Ablations}

\textbf{Setting:} As the number of pixels used for sub-sampling is an important aspect of pose estimation~\cite{iNeRF} with significant implications on runtime, we investigate the effect of pixel sub-sampling on DPPE. Results were computed on 20 randomly-perturbed poses from the Chair, Lego, Materials, and Ship scenes, via the same techniques and evaluation criteria as in Section~\ref{sec:synth}. The percent of pixels randomly sampled from the image includes: 1\%, 25\%, 50\%, 75\%, and 100\%, corresponding to: 6,400, 160,000, 320,000, 480,000, and 640,000 rays, respectively.

\textbf{Results:} Results for pixel sub-sampling are summarized in Table~\ref{tab:table2}. Generally, pose error was similar across all pixel sub-samplings. Using the full image resulted in the best rotation error, and 50\% of the image gave the best translation error. The most significant difference between the various sub-sampling strategies is in the runtime of the algorithm, for which, on average, using 1\% of the image for pose estimation takes 11 seconds, and using the full image takes 73 seconds. As rendering must be performed for $\pm 1 \forall 6$ camera DoF for every 2 epochs (for the rotation step, then translation step), a greater percent of the image rendered for each step results in a greater runtime. The occasional improved performance of DPPE-1\% vs. DPPE-full  is a result of the random selection of pixels helping the algorithm dodge or escape poor local minima by chance. DPPE-1\% is a good candidate for use, due to its reduced runtime, and good performance.

\subsection{Grid Resolution Ablations}

\textbf{Setting:} The resolution of the Plenoxels~\cite{Plenoxels} grid is another important parameter, as higher resolution results in more clear renderings, at the cost of performance. Additionally, as the step-size for translational image gradient estimation was set to $\pm \ 1$ voxel, a lower resolution results in a coarser gradient estimate. Results were computed on 20 randomly-perturbed poses from the Chair, Lego, Materials, and Ship scenes, with each perturbation being random across resolution. Otherwise, results were computed via the same methods as in Section~\ref{sec:synth}. Grid resolutions of $[64]^3, [128]^3, [256]^3,$ and $[512]^3$ were selected, and a Plenoxels grid was trained for each resolution.

\textbf{Results:} The effect of grid resolution on DPPE pose error is presented in Table~\ref{tab:table3}. Rotationally, a grid of $[256]^3$ resulted in the lowest pose error, and translationally a $[512]^3$ grid was the best. Results were significantly different for resolutions of 64 and 128, though 256 and 512 offered relatively similar performance. A greater resolution requires a more costly render, thereby increasing runtime.

Additional informal ablations were conducted for the pose estimation algorithm, including on the optimization alternation strategy (e.g. 1-rotation-1-translation, 100-rotation-100-translation, etc.), and image gradient estimation step-size (e.g. $\pm \ 0.1, 0.01, 0.001$, etc. radians for rotational gradient estimation), and the best-performing strategies were used uniformly for all experiments.

\subsection{Limitations and Future Work}

\begin{figure}[ht]
    \centering
    {\includegraphics[width=0.4\textwidth]{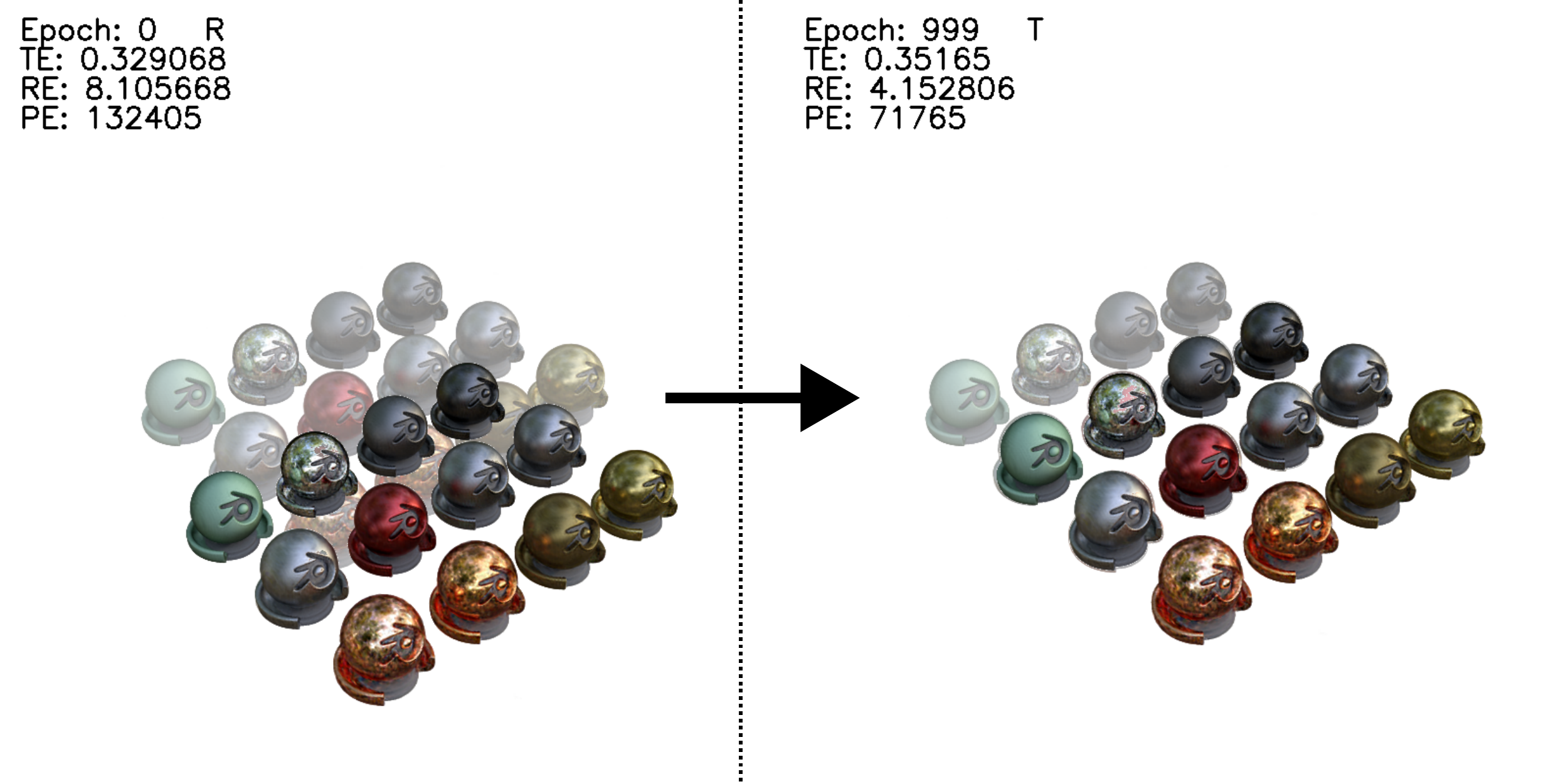}} 
    \caption[Local Minima]{DPPE can be susceptible to local minima. (Left) The scene rendered from the pose initialization (opaque) with the ground-truth (transparent). (Right) The render of the scene from the final pose estimate, after DPPE. As DPPE only considers the local gradient approximation, due to the local minimum the pose is misaligned by an entire row.}
    \label{fig:local_minima}
\end{figure}

One of the most common issues with DPPE's optimization is that it is vulnerable to local minima, especially in certain scenes, such as Materials. An illustration of such a case is presented in Fig.~\ref{fig:local_minima}.
To help overcome this, methods such as Monte Carlo sampling could be employed, as starting with multiple candidate pose initializations has been shown to alleviate the local minima issue for neural rendering pose estimation~\cite{parallel_inv}. Depth data could also be integrated for this, though it would reduce the applicability of the work.

As this method is theoretically invariant of the scene representation technique (other than some implementation details such as the gradient estimation step size), it could be extended to works such as classical NeRF~\cite{NeRF}, Instant NGP~\cite{instantNGP} or 3D Gaussian Splatting~\cite{kerbl3Dgaussians}. An extension to the latter, in particular, might be beneficial, as its rendering is even faster than that of Plenoxels~\cite{kerbl3Dgaussians}.

\section{CONCLUSIONS} \label{sec:conclusion}

In this work, we have introduced DPPE, a 6DoF monocular RGB pose estimation technique that operates over a Plenoxels scene representation. DPPE shows comparable results with existing neural pose estimation works, with a quicker runtime for the DPPE-1\% variant. We demonstrate that the rapid rendering speed of recent neural rendering works can be effectively leveraged to numerically approximate the image gradient via central differencing. Furthermore, we show that this numerical approximation can be used with a modified classical template-matching approach to perform pose estimation. Finally, we analyze the effects of key algorithm parameters, including ray sub-sampling technique and resolution, which indicate that even 1\% of the image is effective for pose estimation. As DPPE can theoretically operate over any neural scene representation, future research will look to integrate it with state-of-the-art neural rendering techniques, as well as methods for reducing the impact of local minima.








\bibliography{references}

\begin{thebibliography}{10}
\providecommand{\url}[1]{#1}
\csname url@samestyle\endcsname
\providecommand{\newblock}{\relax}
\providecommand{\bibinfo}[2]{#2}
\providecommand{\BIBentrySTDinterwordspacing}{\spaceskip=0pt\relax}
\providecommand{\BIBentryALTinterwordstretchfactor}{4}
\providecommand{\BIBentryALTinterwordspacing}{\spaceskip=\fontdimen2\font plus
\BIBentryALTinterwordstretchfactor\fontdimen3\font minus \fontdimen4\font\relax}
\providecommand{\BIBforeignlanguage}[2]{{%
\expandafter\ifx\csname l@#1\endcsname\relax
\typeout{** WARNING: IEEEtran.bst: No hyphenation pattern has been}%
\typeout{** loaded for the language `#1'. Using the pattern for}%
\typeout{** the default language instead.}%
\else
\language=\csname l@#1\endcsname
\fi
#2}}
\providecommand{\BIBdecl}{\relax}
\BIBdecl

\bibitem{Cadena2016TRO}
C.~Cadena, L.~Carlone, H.~Carrillo, Y.~Latif, D.~Scaramuzza, J.~Neira, I.~Reid, and J.~J. Leonard, ``Past, present, and future of simultaneous localization and mapping: Toward the robust-perception age,'' \emph{IEEE Transactions on Robotics}, vol.~32, no.~6, pp. 1309--1332, 2016.

\bibitem{analysisbysynthesis}
X.~Chen, Z.~Dong, J.~Song, A.~Geiger, and O.~Hilliges, ``Category level object pose estimation via neural analysis-by-synthesis,'' in \emph{Computer Vision – ECCV 2020: 16th European Conference, Glasgow, UK, August 23–28, 2020, Proceedings, Part XXVI}.\hskip 1em plus 0.5em minus 0.4em\relax Berlin, Heidelberg: Springer-Verlag, 2020, p. 139–156.

\bibitem{GDRNet}
G.~Wang, F.~Manhardt, F.~Tombari, and X.~Ji, ``{GDR-}net: Geometry-guided direct regression network for monocular 6d object pose estimation,'' in \emph{2021 IEEE/CVF Conference on Computer Vision and Pattern Recognition (CVPR)}.\hskip 1em plus 0.5em minus 0.4em\relax Los Alamitos, CA, USA: IEEE Computer Society, jun 2021, pp. 16\,606--16\,616.

\bibitem{NeRF}
B.~Mildenhall, P.~P. Srinivasan, M.~Tancik, J.~T. Barron, R.~Ramamoorthi, and R.~Ng, ``{NeRF: representing scenes as neural radiance fields for view synthesis},'' \emph{Communications of the ACM}, vol.~65, no.~1, pp. 99--106, 2021.

\bibitem{iNeRF}
L.~Yen-Chen, P.~Florence, J.~T. Barron, A.~Rodriguez, P.~Isola, and T.-Y. Lin, ``{iNeRF: Inverting Neural Radiance Fields for Pose Estimation},'' \emph{2021 IEEE/RSJ International Conference on Intelligent Robots and Systems (IROS)}, vol.~00, pp. 1323--1330, 2021.

\bibitem{BARF}
C.-H. Lin, W.-C. Ma, A.~Torralba, and S.~Lucey, ``{BARF: Bundle-Adjusting Neural Radiance Fields},'' \emph{2021 IEEE/CVF International Conference on Computer Vision (ICCV)}, vol.~00, pp. 5721--5731, 2021.

\bibitem{Plenoxels}
S.~Fridovich-Keil, A.~Yu, M.~Tancik, Q.~Chen, B.~Recht, and A.~Kanazawa, ``{Plenoxels: Radiance Fields without Neural Networks},'' \emph{2022 IEEE/CVF Conference on Computer Vision and Pattern Recognition (CVPR)}, vol.~00, pp. 5491--5500, 2022.

\bibitem{instantNGP}
T.~Müller, A.~Evans, C.~Schied, and A.~Keller, ``{Instant neural graphics primitives with a multiresolution hash encoding},'' \emph{ACM Transactions on Graphics (TOG)}, vol.~41, no.~4, pp. 1--15, 2022.

\bibitem{kerbl3Dgaussians}
B.~Kerbl, G.~Kopanas, T.~Leimk{\"u}hler, and G.~Drettakis, ``3d gaussian splatting for real-time radiance field rendering,'' \emph{ACM Transactions on Graphics}, vol.~42, no.~4, July 2023.

\bibitem{BeyondTrilerp}
B.~Csébfalvi, ``{Beyond trilinear interpolation},'' \emph{ACM Transactions on Graphics (TOG)}, vol.~38, no.~4, pp. 1--8, 2019.

\bibitem{prattimageprocessing}
K.~W. Pratt, \emph{Image Sampling and Reconstruction}.\hskip 1em plus 0.5em minus 0.4em\relax John Wiley \& Sons, Ltd, 2001, ch.~4, pp. 91--120.

\bibitem{lucaskanade}
B.~D. Lucas and T.~Kanade, ``An iterative image registration technique with an application to stereo vision,'' in \emph{Proceedings of the 7th International Joint Conference on Artificial Intelligence - Volume 2}, ser. IJCAI'81.\hskip 1em plus 0.5em minus 0.4em\relax San Francisco, CA, USA: Morgan Kaufmann Publishers Inc., 1981, p. 674–679.

\bibitem{nerffourier}
M.~Tancik, P.~Srinivasan, B.~Mildenhall, S.~Fridovich-Keil, N.~Raghavan, U.~Singhal, R.~Ramamoorthi, J.~Barron, and R.~Ng, ``{Fourier Features Let Networks Learn High Frequency Functions in Low Dimensional Domains},'' in \emph{Advances in Neural Information Processing Systems}, H.~Larochelle, M.~Ranzato, R.~Hadsell, M.~Balcan, and H.~Lin, Eds., vol.~33.\hskip 1em plus 0.5em minus 0.4em\relax Curran Associates, Inc., 2020, pp. 7537--7547.

\bibitem{mipnerf}
J.~T. Barron, B.~Mildenhall, D.~Verbin, P.~P. Srinivasan, and P.~Hedman, ``{Mip-NeRF 360: Unbounded Anti-Aliased Neural Radiance Fields},'' \emph{2022 IEEE/CVF Conference on Computer Vision and Pattern Recognition (CVPR)}, vol.~00, pp. 5460--5469, 2022.

\bibitem{blockNeRF}
M.~Tancik, V.~Casser, X.~Yan, S.~Pradhan, B.~P. Mildenhall, P.~Srinivasan, J.~T. Barron, and H.~Kretzschmar, ``{Block-NeRF: Scalable Large Scene Neural View Synthesis},'' in \emph{2022 IEEE/CVF Conference on Computer Vision and Pattern Recognition (CVPR)}, 2022, pp. 8238--8248.

\bibitem{neuralflow}
Z.~Li, S.~Niklaus, N.~Snavely, and O.~Wang, ``Neural scene flow fields for space-time view synthesis of dynamic scenes,'' in \emph{2021 IEEE/CVF Conference on Computer Vision and Pattern Recognition (CVPR)}, 2021, pp. 6494--6504.

\bibitem{pnp}
L.~Quan and Z.~Lan, ``{Linear N-point camera pose determination},'' \emph{IEEE Transactions on Pattern Analysis and Machine Intelligence}, vol.~21, no.~8, pp. 774--780, 1999.

\bibitem{colmap}
J.~L. Schönberger and J.-M. Frahm, ``{Structure-from-Motion Revisited},'' \emph{2016 IEEE Conference on Computer Vision and Pattern Recognition (CVPR)}, pp. 4104--4113, 2016.

\bibitem{monoslam}
A.~J. Davison, I.~D. Reid, N.~D. Molton, and O.~Stasse, ``{MonoSLAM: Real-Time Single Camera SLAM},'' \emph{IEEE Transactions on Pattern Analysis and Machine Intelligence}, vol.~29, no.~6, pp. 1052--1067, 2007.

\bibitem{orbslam3}
C.~Campos, R.~Elvira, J.~J.~G. Rodrguez, J.~M.~M. Montiel, and J.~D. Tards, ``{ORB-SLAM3: An Accurate Open-Source Library for Visual, VisualInertial, and Multimap SLAM},'' \emph{IEEE Transactions on Robotics}, vol.~37, no.~6, pp. 1874--1890, 2021.

\bibitem{siftslam}
Z.~Dai-xian, ``{Binocular Vision-SLAM Using Improved SIFT Algorithm},'' \emph{2010 2nd International Workshop on Intelligent Systems and Applications}, pp. 1--4, 2010.

\bibitem{kinectfusion}
R.~A. Newcombe, A.~Fitzgibbon, S.~Izadi, O.~Hilliges, D.~Molyneaux, D.~Kim, A.~J. Davison, P.~Kohi, J.~Shotton, and S.~Hodges, ``{KinectFusion: Real-time dense surface mapping and tracking},'' \emph{2011 10th IEEE International Symposium on Mixed and Augmented Reality}, pp. 127--136, 2011.

\bibitem{gradSLAM}
K.~M. Jatavallabhula, S.~Saryazdi, G.~Iyer, and L.~Paull, ``{gradSLAM: Automagically differentiable SLAM},'' \emph{arXiv}, 2019.

\bibitem{cNNopticalflow}
A.~Dosovitskiy, P.~Fischer, E.~Ilg, P.~Hausser, C.~Hazirbas, V.~Golkov, P.~V.~D. Smagt, D.~Cremers, and T.~Brox, ``{FlowNet: Learning Optical Flow with Convolutional Networks},'' \emph{2015 IEEE International Conference on Computer Vision (ICCV)}, pp. 2758--2766, 2015.

\bibitem{NeRF--}
Z.~Wang, S.~Wu, W.~Xie, M.~Chen, and V.~A. Prisacariu, ``{NeRF--: Neural Radiance Fields Without Known Camera Parameters},'' \emph{arXiv}, 2021.

\bibitem{parallel_inv}
Y.~Lin, T.~Müller, J.~Tremblay, B.~Wen, S.~Tyree, A.~Evans, P.~A. Vela, and S.~Birchfield, ``{Parallel Inversion of Neural Radiance Fields for Robust Pose Estimation},'' \emph{2023 IEEE International Conference on Robotics and Automation (ICRA)}, vol.~00, pp. 9377--9384, 2023.

\bibitem{locnerf}
D.~Maggio, M.~Abate, J.~Shi, C.~Mario, and L.~Carlone, ``Loc-nerf: Monte carlo localization using neural radiance fields,'' in \emph{2023 IEEE International Conference on Robotics and Automation (ICRA)}, 2023, pp. 4018--4025.

\bibitem{iMAP}
E.~Sucar, S.~Liu, J.~Ortiz, and A.~J. Davison, ``{iMAP: Implicit Mapping and Positioning in Real-Time},'' \emph{2021 IEEE/CVF International Conference on Computer Vision (ICCV)}, vol.~00, pp. 6209--6218, 2021.

\bibitem{NICE-SLAM}
Z.~Zhu, S.~Peng, V.~Larsson, W.~Xu, H.~Bao, Z.~Cui, M.~R. Oswald, and M.~Pollefeys, ``{NICE-SLAM: Neural Implicit Scalable Encoding for SLAM},'' \emph{2022 IEEE/CVF Conference on Computer Vision and Pattern Recognition (CVPR)}, vol.~00, pp. 12\,776--12\,786, 2022.

\bibitem{plenoxelslam}
A.~L. Teigen, Y.~Park, A.~Stahl, and R.~Mester, ``{RGB-D Mapping and Tracking in a Plenoxel Radiance Field},'' \emph{IEEE/CVF WACV}, pp. 3342--–3351, 2024.

\bibitem{LiDARSLAM}
A.~Koval, C.~Kanellakis, and G.~Nikolakopoulos, ``Evaluation of lidar-based 3d slam algorithms in subt environment,'' \emph{IFAC-PapersOnLine}, vol.~55, no.~38, pp. 126--131, 2022, 13th IFAC Symposium on Robot Control SYROCO 2022.

\bibitem{orbeezslam}
C.-M. Chung, Y.-C. Tseng, Y.-C. Hsu, X.-Q. Shi, Y.-H. Hua, J.-F. Yeh, W.-C. Chen, Y.-T. Chen, and W.~H. Hsu, ``{Orbeez-SLAM: A Real-time Monocular Visual SLAM with ORB Features and NeRF-realized Mapping},'' \emph{2023 IEEE International Conference on Robotics and Automation (ICRA)}, vol.~00, pp. 9400--9406, 2023.

\bibitem{NeRF-SLAM}
A.~Rosinol, J.~J. Leonard, and L.~Carlone, ``{NeRF-SLAM: Real-Time Dense Monocular SLAM with Neural Radiance Fields},'' \emph{arXiv}, 2022.

\bibitem{Plenoctrees}
A.~Yu, R.~Li, M.~Tancik, H.~Li, R.~Ng, and A.~Kanazawa, ``{PlenOctrees for Real-time Rendering of Neural Radiance Fields},'' \emph{2021 IEEE/CVF International Conference on Computer Vision (ICCV)}, vol.~00, pp. 5732--5741, 2021.

\bibitem{microLie}
J.~Solà, J.~Deray, and D.~Atchuthan, ``{A micro Lie theory for state estimation in robotics},'' \emph{arXiv}, 2018.

\bibitem{Blender}
B.~O. Community, \emph{Blender - a 3D modelling and rendering package}, Blender Foundation, Stichting Blender Foundation, Amsterdam, 2018.

\end{thebibliography}



\end{document}